\documentclass{article} 
\usepackage{iclr2026_conference,times}
\usepackage{graphicx}

\usepackage{amsmath,amsfonts,bm}

\def\eqref#1{equation~\ref{#1}}

\def\1{\bm{1}}

\DeclareMathAlphabet{\mathsfit}{\encodingdefault}{\sfdefault}{m}{sl}
\SetMathAlphabet{\mathsfit}{bold}{\encodingdefault}{\sfdefault}{bx}{n}

\usepackage{hyperref}
\usepackage{url}

\title{
Agentified Assessment of Logical Reasoning Agents
}

\author{Zhiyu Ni\thanks{Equal contribution.} \\
University of California, Berkeley \\
zhiyuni@berkeley.edu
\And
Yifeng Xiao\footnotemark[1] \\
University of California, Berkeley \\
yifeng\_xiao@berkeley.edu
\And
Zheng Liang\footnotemark[1] \\
University of California, Berkeley \\
zhliang@berkeley.edu
}

\iclrfinalcopy 

\begin{document}

\maketitle

\begin{abstract}
We present a framework for evaluating and benchmarking logical reasoning agents when assessment itself must be reproducible, auditable, and robust to execution failures.
Building on agentified assessment, we use an assessor agent to issue tasks, enforce execution budgets, parse outputs, and record structured failure types, while the agent under test only needs to expose a standardized agent-to-agent interface. 
As a case study, we benchmark an auto-formalization agent for first-order logic (FOL) reasoning on a solver-verified and repaired split of FOLIO. The agent translates natural language premises and conclusions into executable Z3Py programs and employs satisfiability modulo theories (SMT) solving to determine logical entailment.
On the cleaned FOLIO validation set, the auto-formalization agent achieves $86.70\%$ accuracy under the assessor protocol, outperforming a chain-of-thought baseline ($73.89\%$).

\end{abstract}

\section{Introduction}
\label{sec:intro}



Evaluating and benchmarking reasoning agents is challenging because failures occur at multiple layers, including model reasoning and tool execution.
As a result, static evaluation harnesses often conflate operational failures (e.g., timeouts, runtime errors, output parsing failures) with reasoning errors, and may hide these failure modes behind a single accuracy number.
Moreover, traditional setups tightly couple benchmark logic to agent implementations, so integration effort grows with the number of benchmarks.

In this work, we study first-order logic (FOL) inference on FOLIO~\citep{han2024folio} and adopt agentified agent assessment (AAA) as described in AgentBeats~\citep{agentbeats2026aaa}, where assessment logic is implemented as an assessor agent that interacts with an agent under test through a standardized agent-to-agent (A2A) interface~\citep{a2a2026spec}.

We conduct verification and repair on the FOLIO dataset~\citep{han2024folio} using a data cleaning pipeline, establishing a more reliable benchmark.
On the cleaned benchmark, we implement a {white-box auto-formalization agent} that achieves {86.70\%} accuracy on the FOLIO validation set, outperforming a chain-of-thought baseline ({73.89\%})~\citep{wei2022chain}. 


\section{Benchmark and Data Cleaning}
\label{sec:benchmark}

We evaluate on {FOLIO}~\citep{han2024folio}, a first-order logic (FOL) reasoning benchmark consisting of natural-language premises and conclusions paired with formal FOL annotations. FOLIO is derived from Wikipedia-based real-world scenarios and requires both natural language understanding and logical reasoning.
For each example, FOLIO provides one of three labels: \textsc{True} (the conclusion is logically entailed by the premises), \textsc{False} (the conclusion contradicts the premises), or \textsc{Uncertain} (the conclusion cannot be determined from the premises). We evaluate classification accuracy against human-annotated ground-truth labels.

\subsection{Verification and Repair Pipeline}

\begin{figure}[t!]
    \centering
    \includegraphics[width=0.85\textwidth]{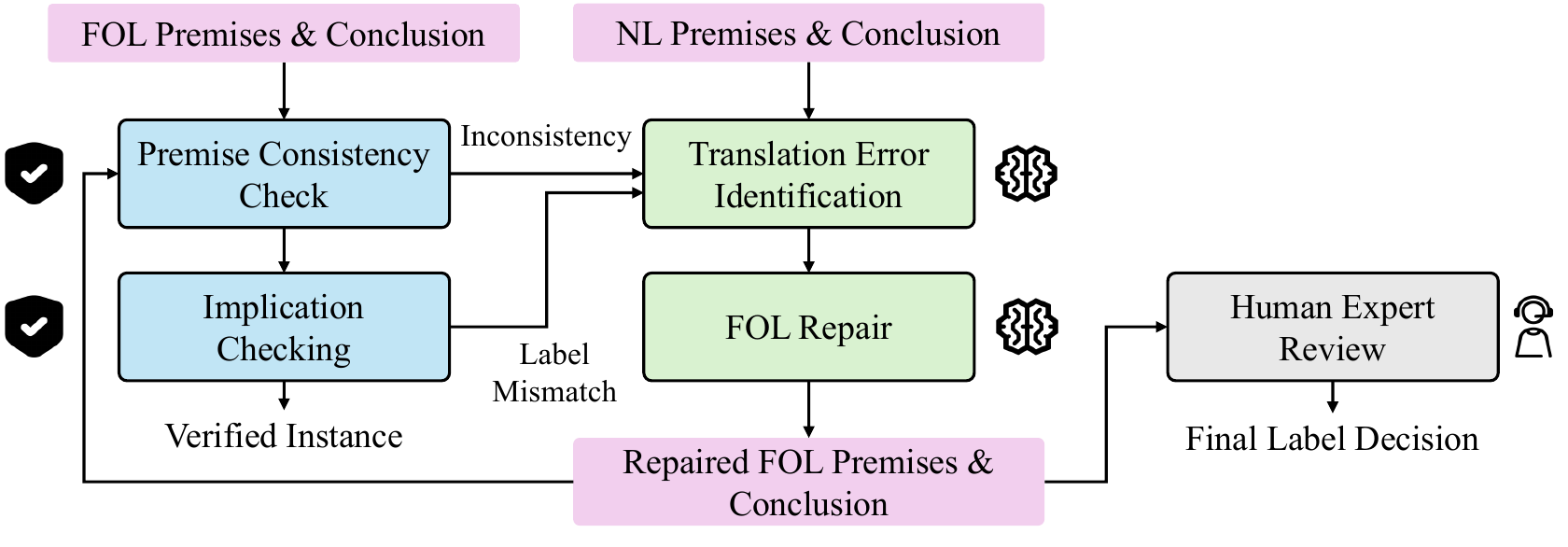}
    \caption{Overview of the data cleaning pipeline.}
    \label{fig:cleaning-pipeline}
\end{figure}

The original FOLIO dataset exhibits potential label errors and misalignments between natural language and formal annotations due to the complexity of semantic parsing. To establish a more reliable benchmark, we implement a systematic data cleaning pipeline (Figure~\ref{fig:cleaning-pipeline}) that leverages symbolic verification using the Vampire theorem prover~\citep{kovacs2013first} to execute formal reasoning on FOL representations. When verification results conflict with expected labels, we identify and iteratively repair translation errors.

Given a premise set $\{\phi_1, \phi_2, \dots \}$ and a conclusion $\varphi$, we detail the pipeline as follows:

\noindent \textbf{Formal Premise-Conclusion Checking}. 
We verify the implication relation between premises and conclusion in two steps. First, we check whether the premises are consistent (i.e., whether $\bigwedge_i \phi_i$ is satisfiable). Then, we perform implication checking using the following satisfiability conditions:
\begin{align}
\textsc{True} &\iff \bigwedge_{i} \phi_i \rightarrow \varphi \iff \bigwedge_{i} \phi_i \wedge \neg \varphi \text{ is unsatisfiable}, \\
\textsc{False} &\iff \bigwedge_{i} \phi_i \rightarrow \neg \varphi \iff \bigwedge_{i} \phi_i \wedge \varphi \text{ is unsatisfiable}.
\end{align}
When both implication checks fail, we label the instance as \textsc{Uncertain}. We then compare verification results against expected labels to identify potential annotation errors.


\noindent \textbf{NL-FOL Misalignment Identification and Repair.}
When premises are inconsistent or verification results conflict with expected labels, we employ a check-and-fix procedure using two LLM-based agents. 
A \textit{critique agent} diagnoses systematic translation errors (e.g., unbalanced parentheses, lexical typos, and naming convention inconsistencies) by analyzing the original natural language and FOL annotations.
A \textit{refiner agent} then executes targeted corrections. We iteratively re-verify the repaired FOL until the expected labels are achieved. 
If the number of iterations exceeds a predefined threshold without resolution, we flag the instance for manual review. We release the cleaned and repaired FOLIO split at \url{https://huggingface.co/datasets/yfxiao/folio-refined}.

\subsection{Benchmark Statistics}

We classify instances into three categories: (i) verified directly with the original FOL, (ii) verified after automatic repair, and (iii) problematic instances flagged for manual review.
For the \textit{training set} (1,001 examples): 674 ($67.3\%$) verified directly, 23 ($2.3\%$) after automatic repair, and 304 ($30.4\%$) remain problematic.
For the \textit{validation set} (203 examples): 154 ($75.9\%$) verified directly, 10 ($4.9\%$) after automatic repair, and 39 ($19.2\%$) remain problematic.
In total, we identify $3.8\%$ potential label errors in the training set and $1.5\%$ in the validation set, improving both the NL-FOL translation quality and the reliability of labels for evaluation. 

\section{Agentified Evaluation Framework}
\label{sec:methodology}

\subsection{Agentified Assessment}
\label{sec:agentified-assessment}

We formalize the evaluation of reasoning agents by \emph{treating assessment itself as an agent}.
Instead of relying on a static evaluation script, we separate the system into two interacting components: an \emph{agent under test}, which performs reasoning, and an \emph{assessor agent}, which controls task execution, interprets outcomes, and assigns scores.
We adopt the agentified agent assessment (AAA) abstraction described in AgentBeats~\citep{agentbeats2026aaa}, where the assessor and the agent under test communicate through an agent-to-agent (A2A) interface~\citep{a2a2026spec}.

The primary value of AAA is not to address individual operational issues (e.g., timeouts), which can also be handled in traditional benchmarks.
Instead, it changes the integration cost model for benchmarking agents: in traditional setups, integration cost grows with the number of benchmarks ($O(n)$), whereas under AAA, an agent implements A2A once and can participate in many assessors ($O(1)$).
This decoupling enables plug-and-play evaluation and architecture freedom for agents with diverse internal designs.

\begin{figure}[t!]
    \centering
    \includegraphics[width=0.65\linewidth]{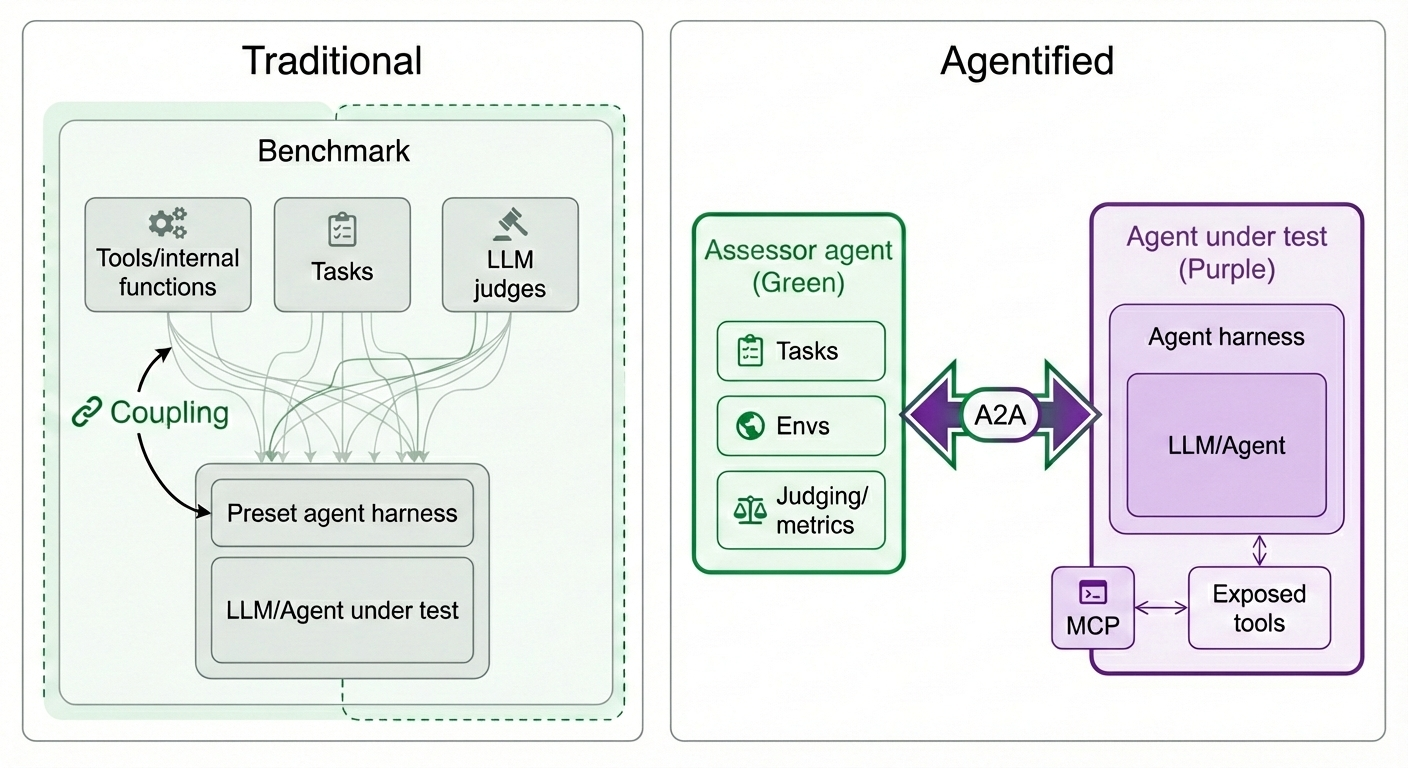}
    \caption{Traditional evaluation harnesses couple task execution, environments, and judging logic to a preset harness. In agentified assessment, an assessor agent evaluates an agent under test via an A2A interface~\citep{a2a2026spec}, reducing integration overhead.}
    \label{fig:agentified-assessment}
\end{figure}

As illustrated in Figure~\ref{fig:agentified-assessment}, traditional benchmarks couple task orchestration and judging to a benchmark-specific harness.
AAA decouples these components by packaging evaluation logic as an assessor agent that interacts with an agent under test through a standardized interface, allowing the assessor to evolve without breaking participants.

\subsection{Assessor Agent (Evaluation Protocol)}
\label{sec:assessor-agent}

The \emph{assessor agent} implements the evaluation protocol.
For each premise and conclusion pair, it issues the task to the agent under test, enforces execution budgets (e.g., timeouts and retries), parses and validates the response, and assigns a final label (\textsc{True}, \textsc{False}, or \textsc{Uncertain}) according to the benchmark criteria.

If the agent under test outputs additional text, the assessor parses the final label deterministically; non-parseable outputs are recorded as \textsc{ParseError}.
Rather than discarding failures, the assessor records structured failure types such as \textsc{Timeout}, \textsc{RuntimeError}, and \textsc{ParseError}.
Finally, the assessor emits a machine-consumable evaluation artifact containing per-instance records (gold label, predicted label, correctness, error type if any, latency) and aggregate metrics.

\subsection{Reasoning Agents Under Test}
\label{sec:agent-under-test}

We benchmark two agents under the same assessor.

\noindent \textbf{Chain-of-thought baseline.}
We use chain-of-thought prompting~\citep{wei2022chain}: the agent is instructed to reason step by step and then output its final answer as the last line containing exactly one label in \{\textsc{True}, \textsc{False}, \textsc{Uncertain}\}, which is parsed by the assessor.

\noindent \textbf{Auto-formalization agent.}
The auto-formalization agent solves first-order logic (FOL) inference problems by translating natural language premises and conclusions into executable symbolic programs and performing logical reasoning via solver execution~\citep{pan2023logic, ni2026draft}.
The reasoning process follows a two-stage pipeline.
In Stage~1 (Code Generation), a language model generates executable Z3Py~\citep{de2008z3} code from the input.
In Stage~2 (Execution and Verification), the generated program is executed in a sandbox environment with a 60-second timeout.

Logical validity is determined through satisfiability checking as specified in the benchmark definition (Section~\ref{sec:benchmark}):
if $\bigwedge_i \phi_i \wedge \neg \varphi$ is unsatisfiable, the conclusion is labeled \textsc{True};
if $\bigwedge_i \phi_i \wedge \varphi$ is unsatisfiable, it is labeled \textsc{False};
otherwise, the label is \textsc{Uncertain}.

To improve robustness, the agent incorporates a self-repair loop with up to three attempts.
When execution fails due to syntax errors or malformed quantifiers, the agent extracts the error message and performs targeted code repair before re-execution.

\paragraph{Implementation note.}
We implement the assessor and agents under test on top of AgentBeats~\citep{agentbeats2026aaa}, where they communicate via A2A~\citep{a2a2026spec} (and optionally expose tools via MCP~\citep{anthropic2024mcp}).
We also constructed a logical reasoning leaderboard that runs the assessor against registered agents and records per-run artifacts (accuracy, latency, and failure types), enabling reproducible comparisons across agents.

\section{Experiments}
\label{sec:experiments}

\subsection{Experimental Setup and Baseline}

We evaluate on the cleaned FOLIO validation set comprising 203 examples after label verification.
Both the chain-of-thought baseline agent and the auto-formalization agent use Gemini 2.5 Flash~\citep{comanici2025gemini} as the backbone LLM with a temperature $T=0.0$ for deterministic outputs and default maximum output tokens $n=65535$. Both agents use a system-user prompt structure. For the chain-of-thought baseline, the system prompt elicits step-by-step reasoning and requires the final answer label to appear as the last line; for auto-formalization, it includes Z3 code generation instructions with syntax guidelines. No in-context learning (ICL) examples are provided for either agent.

\subsection{Results and Analysis}

Table~\ref{tab:breakdown} reports accuracy broken down by ground-truth label category, together with overall performance.
The chain-of-thought baseline achieves an overall accuracy of $73.89\%$ (150/203), while the auto-formalization agent improves accuracy to $86.70\%$ (176/203).

\begin{table}[h!]
\centering
\begin{tabular}{lccc|ccc}
\hline
& \multicolumn{3}{c|}{\textbf{Chain-of-Thought}} & \multicolumn{3}{c}{\textbf{Auto-formalization}} \\
\textbf{Category} & Correct & Total & Acc. & Correct & Total & Acc. \\
\hline
True & 65 & 73 & 89.04\% & 66 & 73 & 90.41\% \\
False & 27 & 61 & 44.26\% & 47 & 61 & 77.05\% \\
Uncertain & 58 & 69 & 84.06\% & 63 & 69 & 91.30\% \\
\hline
\textbf{Overall} & 150 & 203 & 73.89\% & 176 & 203 & 86.70\% \\
\hline
\end{tabular}
\caption{Per-category and overall accuracy on the cleaned FOLIO validation set.}
\label{tab:breakdown}
\end{table}

The largest improvement is observed in the \textsc{False} category, where accuracy increases from $44.26\%$ to $77.05\%$.
Performance on \textsc{True} cases is comparable across methods.
The auto-formalization agent also improves on \textsc{Uncertain} cases ($84.06\%$ to $91.30\%$), highlighting the advantage of solver-based reasoning in handling logical indeterminacy.
Overall, these results demonstrate that formal verification improves robustness on this benchmark, leading to higher overall accuracy.

\section{Conclusion}

This work benchmarks logical reasoning agents under agentified assessment, where an assessor agent enforces budgets, records structured failures, and emits auditable evaluation artifacts. As a case study, an auto-formalization agent that translates natural language problems into executable Z3Py code and uses solver execution outperforms a chain-of-thought baseline on a solver-verified and repaired split of FOLIO, with the largest gains on contradiction (\textsc{False}) cases and consistent improvements on indeterminate (\textsc{Uncertain}) cases. Future work could extend assessor policies and apply agentified assessment to richer tool-using agent settings.

\bibliography{iclr2026_conference}
\bibliographystyle{iclr2026_conference}


\end{document}